\documentclass[runningheads]{llncs}
\usepackage{graphicx}
\usepackage{amsmath,amssymb} % define this before the line numbering.
\usepackage{color}
\usepackage[width=122mm,left=12mm,paperwidth=146mm,height=193mm,top=12mm,paperheight=217mm]{geometry}

\usepackage{caption} 
\usepackage{epsfig}
\usepackage{multirow} 
\graphicspath{ {figures/} }
\usepackage{xspace}
\usepackage{nicefrac}
\usepackage{authblk}
\usepackage[symbol*]{footmisc}

\newcommand{\latinphrase}[1]{\textit{#1}}  
\newcommand{\etal}{\latinphrase{et~al.}\xspace}
\newcommand{\ie}{\latinphrase{i.e.}\xspace}
\newcommand{\eg}{\latinphrase{e.g.}\xspace}
\newcommand{\wrt}{\latinphrase{w.r.t.}\xspace}

\begin{document}
% \linenumbers
\pagestyle{headings}
\mainmatter

\title{Learning and Matching Multi-View Descriptors for Registration of Point Clouds} 
% Replace with your title

\titlerunning{Multi-View Descriptor and Robust Matching }
% Replace with a meaningful short version of your title

\authorrunning{Lei Zhou, Siyu Zhu, Zixin Luo \etal}
% Replace with shorter version of the author list. If there are more authors than fits a line, please use A. Author et al.

\author{Lei Zhou$^1$, Siyu Zhu$^1$\thanks{Siyu Zhu is the corresponding author of this paper.}, Zixin Luo$^1$, Tianwei Shen$^1$, \\Runze Zhang$^1$\thanks{Runze Zhang is the corresponding author.}, Mingmin Zhen$^1$, Tian Fang$^2$\thanks{Tian Fang is with Shenzhen Zhuke Innovation Technology since 2017.}, Long Quan$^1$}

%Please write out author names in full in the paper, i.e. full given and family names. 
%If any authors have names that can be parsed into FirstName LastName in multiple ways, please include the correct parsing, in a comment to the volume editors:
%\index{Lastnames, Firstnames}
%(Do not uncomment it, because you may introduce extra index items if you do that, we will use scripts for introducing index entries...)

\institute{
$^1$~Hong Kong University of Science and Technology, \\
\email{ \{lzhouai,szhu,zluoag,tshenaa,rzhangaj,mzhen,quan\}@cse.ust.hk} \\
$^2$~Altizure.com, \\
\email{ fangtian@altizure.com}
}

\maketitle

\begin{abstract}
Critical to the registration of point clouds is the establishment
of a set of accurate correspondences between points in 3D space. 
The correspondence problem is generally addressed by the design of discriminative 3D local descriptors on the one hand, 
and the development of robust matching strategies on the other hand. 
In this work, we first propose a multi-view local descriptor, which is learned from the images of multiple
views, for the description of 3D keypoints. 
Then, we develop a robust matching
approach, aiming at rejecting outlier matches based on the efficient
inference via belief propagation on the defined graphical model.
We have demonstrated the boost of our approaches to registration on the public scanning and multi-view stereo datasets.
The superior performance has been verified by the intensive comparisons
against a variety of descriptors and matching methods.

\keywords{Point cloud registration; 3D descriptor; Robust matching}
\end{abstract}

\section{Introduction}
Registration of point clouds integrates 3D data from different sources into a common coordinate system,
serving as an essential component of many high-level applications like 3D modeling \cite{Zhu_2018_CVPR,zhang2018distributed}, SLAM \cite{dissanayake2001solution} and robotic perception \cite{fraundorfer2012visual}. 
Critical to a registration task is the determination of correspondences between spatially localized 3D points within each cloud.
To tackle the correspondence problem, on the one hand, a bunch of 3D local descriptors \cite{johnson1999using,rusu2008aligning,rusu2009fast,tombari2010SHOT,tombari2010USC,zeng20163dmatch,Khoury_2017_ICCV,deng2018ppfnet} have been developed to facilitate the description of 3D keypoints. 
On the other hand, matching strategies \cite{lowe2004distinctive,gold1995new,zhou2016fast} have also been progressing towards higher accuracy and robustness.

The exploration of 3D geometric descriptors has long been the focus of interest in point cloud registration. 
It involves the hand-crafted geometric descriptors \cite{johnson1999using,rusu2008aligning,rusu2009fast,tombari2010SHOT,tombari2010USC} as well as the learned ones \cite{zeng20163dmatch,Khoury_2017_ICCV,deng2018ppfnet,2018arXiv180207869G}. Both kinds of methods mainly rely on 3D local geometries.
Meanwhile, with the significant progress of CNN-based 2D patch descriptors \cite{simo2015discriminative,tian2017l2,han2015matchnet,yi2016lift,balntas2016learning},
more importance is attached to leveraging the 2D projections for the description of underlying 3D structures \cite{wu20083d,chu2011multi,dai2017bundlefusion,endres20143}. Particularly for the point cloud data generally co-registered with camera images \cite{Matterport3D,handa:etal:ICRA2014,sturm2012benchmark,SSS:2006}, the fusion of multiple image views, which has reported success on various tasks \cite{su2009learning,Su_2015_ICCV,wangdominant,qi2016volumetric}, is expected to further improve the discriminative power of 3D local descriptors.
With this motivation, we propose a multi-view descriptor, named MVDesc, for the description of 3D keypoints based on the synergy of the multi-view fusion techniques and patch descriptor learning.
Rather than a replacement, the MVDesc is well complementary to existing geometric descriptors \cite{johnson1999using,rusu2008aligning,rusu2009fast,tombari2010SHOT,tombari2010USC,zeng20163dmatch,Khoury_2017_ICCV,deng2018ppfnet}.

Given the local descriptors, the matching problem is another vital issue in point cloud registration. 
A set of outlier-free point matches is desired by most registration algorithms \cite{zhou2016fast,besl1992method,pomerleau2013comparing,Rusinkiewicz:2001:EVO,yang2016go,briales2017convex}.
Currently, the matching strategies, \eg, the nearest neighbor search, mutual best \cite{zhou2016fast} and ratio test \cite{lowe2004distinctive}, basically estimate correspondences according to the similarities of local descriptors alone. 
Without considering the global geometric consistency, these methods are prone to spurious matches between locally-similar 3D structures.
Efforts are also spent on jointly solving the outlier suppression via line process and the optimization of global registration \cite{zhou2016fast}. But the spatial organizations of 3D point matches are still overlooked when identifying outliers.
To address this, we develop a robust matching approach by explicitly considering the spatial consistency of point matches in 3D space.
We seek to filter outlier matches based on a graphical model describing their spatial properties and provide an efficient solution via belief propagation.

The main contributions of this work can be summarized twofold. 
1) We are the first to leverage the fusion of multiple image views for the description of 3D keypoints when tackling point cloud registration. 
2) The proposed effective and efficient outlier filter, which is based on a graphical model and solved by belief propagation, remarkably enhances the robustness of 3D point matching.

\section{Related Works}
\smallskip\noindent\textbf{3D local descriptor.}
The representation of a 3D local structure used to rely on 
traditional geometric descriptors such as Spin Images \cite{johnson1999using}, PFH \cite{rusu2008aligning}, FPFH \cite{rusu2009fast}, SHOT \cite{tombari2010SHOT}, USC \cite{tombari2010USC} and \etal, which are mainly produced based on the histograms over local geometric attributes. 
Recent studies seek to learn descriptors from different representations of local geometries, like volumetric representations of 3D patches \cite{zeng20163dmatch}, point sets \cite{deng2018ppfnet} and depth maps \cite{2018arXiv180207869G}.
The CGF \cite{Khoury_2017_ICCV} still leverages the traditional spherical histograms to capture the local geometry but learns to map the high-dimensional histograms to a low-dimensional space for compactness. 

Rather than only using geometric properties, some existing works refer to extracting descriptors from RGB images that are commonly co-registered with point clouds as in scanning datasets \cite{Matterport3D,handa:etal:ICRA2014,sturm2012benchmark} and 3D reconstruction datasets \cite{SSS:2006,strecha2008benchmarking}.  
Registration frameworks like \cite{wu20083d,chu2011multi,dai2017bundlefusion,endres20143} use SIFT descriptors \cite{lowe2004distinctive} as the representations of 3D keypoints based on their projections in single-view RGB images. Besides, the other state-of-the-art 2D descriptors like DeepDesc \cite{simo2015discriminative}, L2-Net \cite{tian2017l2} and \etal \cite{han2015matchnet,yi2016lift,balntas2016learning} can easily migrate here for the description of 3D local structures.

\smallskip\noindent\textbf{Multi-view fusion.}
The multi-view fusion technique is used to integrate information from multiple views into a single representation.
It has been widely proved by the literature that the technique effectively boosts the performance of instance-level detection \cite{su2009learning}, recognition \cite{Su_2015_ICCV,wangdominant} and classification \cite{qi2016volumetric} compared with a single view.
Su \etal \cite{su2009learning} first propose a probabilistic representation of a 3D-object class model for the scenario where an object is positioned at the center of a dense viewing sphere.
A more general strategy of multi-view fusion is \textit{view pooling} \cite{Su_2015_ICCV,wangdominant,qi2016volumetric,chen2016multi}, which aggregates the feature maps of multiple views via element-wise maximum operation.

\smallskip\noindent\textbf{Matching.}
The goal of matching in point cloud registration is to find correspondences across 3D point sets given keypoint descriptors.
Almost all the registration algorithms \cite{zhou2016fast,besl1992method,pomerleau2013comparing,Rusinkiewicz:2001:EVO,yang2016go,briales2017convex} demand accurate point correspondences as input.
Nearest-neighbor search, mutual best filtering \cite{zhou2016fast} and ratio test \cite{lowe2004distinctive} are effective ways of searching for potential matches based on local similarities for general matching tasks.
However, as mentioned above, these strategies are prone to mismatches without considering the geometric consistency.
To absorb geometric information, 
\cite{albarelli2009matching} and \cite{rodola2013scale} discover matches in geometric agreement using game-theoretic scheme.
Ma \etal \cite{ijcai2017-627} propose to reject outliers by enforcing consistency in local neighborhood.
Zhou \etal \cite{zhou2016fast} use a RANSAC-style tuple test to eliminate matches with inconsistent scales. 
Besides, the line process model \cite{black1996unification} is applied in registration domain to account for the presence of outliers implicitly \cite{zhou2016fast}.
\iffalse
\smallskip\noindent\textbf{Registration algorithms.}
Generally, current registration algorithms can be split into two branches.
The first branch uses the ICP and its variants \cite{besl1992method,pomerleau2013comparing,Rusinkiewicz:2001:EVO} which are known to be susceptible to local minima.
To address the local minima problem, another branch explores global approaches.
The Go-ICP \cite{yang2016go} guarantees optimal solutions theoretically based on the branch-and-bound scheme.
Zhou \etal  \cite{zhou2016fast} solve the non-convex registration problem globally via graduated non-convexity. 
Briales \etal  \cite{briales2017convex} address finding the globally optimal transformation by introducing Lagrangian dual relaxation.
It is noteworthy that all these registration algorithms require accurate correspondences of 3D points as input. Therefore, the proposed MVDesc descriptor and robust matching approach serve as a complement to the existing registration algorithms.
\fi

\section{Multi-View Local Descriptor (MVDesc)} \label{sec:descriptor}
In this section, we propose to learn multi-view descriptors (MVDesc) for 3D keypoints which combine multi-view fusion techniques \cite{su2009learning,Su_2015_ICCV,wangdominant,qi2016volumetric} with patch descriptor learning \cite{simo2015discriminative,tian2017l2,han2015matchnet,yi2016lift,balntas2016learning}.
Specifically, we first propose a new view-fusion architecture to integrate feature maps across views into a single representation.
Second, we build the MVDesc network for learning by putting the fusion architecture above multiple feature networks \cite{Han_2015_CVPR}. Each feature network is used to extract feature maps from the local patch of each view.

\begin{figure}[t]
\begin{center}
\includegraphics[width=1\linewidth]{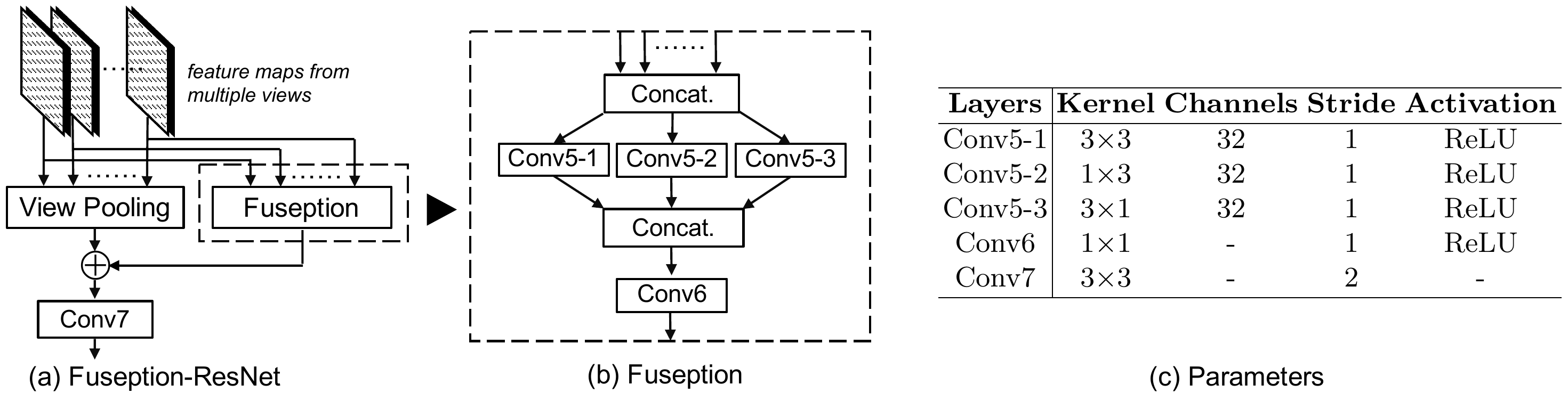}
\end{center}
   \caption{An overview of proposed Fuseption-ResNet (FRN).
   (a) Architecture of FRN that fuses feature maps of multiple views. Backed by the view pooling branch as a shortcut connection, (b) the Fuseption branch takes charge of learning the residual mapping. (c) The parameters of the convolutional layers are listed}
\label{fig:network}
\end{figure}

\iffalse
\begin{table}[]
\centering
\caption{My caption}
\label{my-label}
\begin{tabular}{c|cccc}
\hline
\textbf{Layers}  & \textbf{Kernel}     & \textbf{Channels} & \textbf{Stride} & \textbf{Activation} \\ \hline
Conv5-1 & 3$\times$3 & 32       & 1      & ReLU       \\
Conv5-2 & 1$\times$3 & 32       & 1      & ReLU       \\
Conv5-3 & 3$\times$1 & 32       & 1      & ReLU       \\
Conv6   & 1$\times$1 & -       & 1      & ReLU       \\
Conv7   & 3$\times$3 & -        & 2      & -       \\ \hline
\end{tabular}
\end{table}
\fi

\subsection{Multi-View Fusion}
Currently, \textit{view pooling} is the dominant fusion technique used to merge feature maps from different views \cite{Su_2015_ICCV,wangdominant,qi2016volumetric,chen2016multi}.
However, as reported by the literature \cite{wangdominant,HardNet2017,johnson2016perceptual}, the pooling operation is somewhat risky in terms of feature aggregation due to its effect of smoothing out the subtle local patterns. 
Inspired by ResNet \cite{he2016deep}, we propose an architecture termed \textit{Fuseption-ResNet} which uses the view pooling as a shortcut connection and adds a sibling branch termed \textit{Fuseption} in charge of learning the underlying residual mapping.

\smallskip\noindent\textbf{Fuseption.}
As shown in Figure \ref{fig:network}(b), the Fuseption is an Inception-style \cite{szegedy2015going,szegedy2016rethinking} architecture capped above multi-view feature maps. 
First, following the structure of inception modules \cite{szegedy2015going,szegedy2016rethinking}, three lightweight cross-spatial filters with different kernel sizes, $3\times3$, $1\times3$ and $3\times1$, are adopted to extract different types of features.
Second, the $1\times1$ convolution Conv6, employed above concatenated feature maps, is responsible for the merging of correlation statistics across channels and the dimension reduction as suggested by \cite{szegedy2015going,lin2013network}. 

\smallskip\noindent\textbf{Fuseption-ResNet (FRN).}
Inspired by the effectiveness of skip connections in ResNet \cite{he2016deep}, we take view pooling as a shortcut in addition to Fuseption as shown in Figure \ref{fig:network}(a).
As opposed to the view pooling branch which is in charge of extracting the strongest responses across views \cite{Su_2015_ICCV}, the Fuseption branch is responsible for learning the underlying residual mapping.
Both engaged branches reinforce each other in term of accuracy and convergence rate.
On the one hand, the residual branch, Fuseption, guarantees no worse accuracy compared to just using view pooling.
This is because if view pooling is optimal, the residual can be easily pulled to zeros during training. 
On the other hand, the shortcut branch, view pooling, greatly accelerates the convergence of learning MVDesc as illustrated in Figure \ref{fig:loss}(a).
Intuitively, since the view pooling branch has extracted the essential strongest responses across views, it is easier for the Fuseption branch to just learn the residual mapping.

\subsection{Learning MVDesc}
\begin{figure}[t]
\begin{center}
\includegraphics[width=1.0\linewidth]{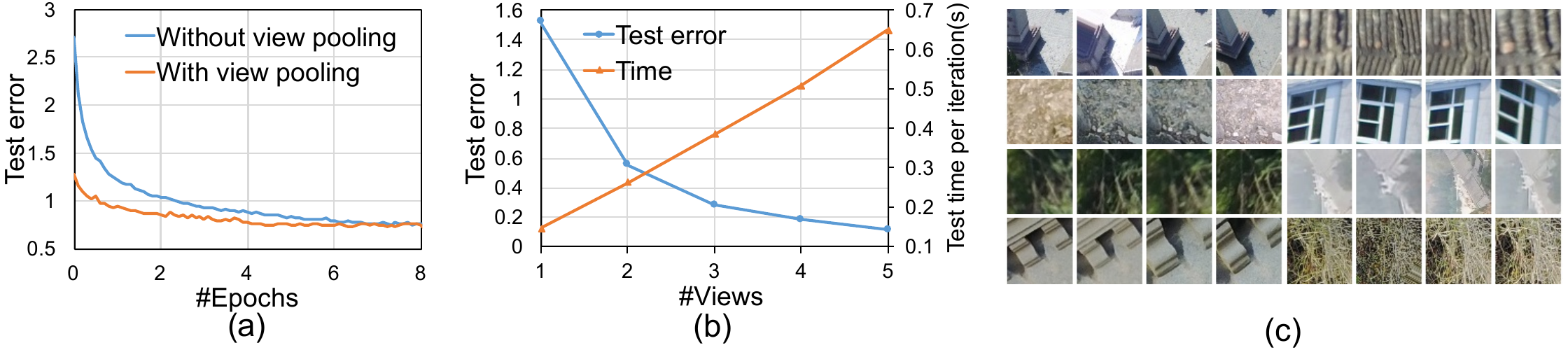}
\end{center}
   \caption{
   (a) Test error of our MVDesc network with or without the view pooling branch. Using view pooling as a shortcut connection contributes to much faster convergence of learning.
   (b) Test error and forward time per iteration of MVDesc network with respect to the number of fused views. 
Three views are chosen as a good trade-off between accuracy and efficiency.
   (c) The sample multi-view patches produced by the collected SfM database for training}
\label{fig:loss}
\end{figure}

\smallskip\noindent\textbf{Network.} The network for learning MVDesc is built by putting the proposed FRN above multiple parallel feature networks.
We use the feature network from MatchNet \cite{Han_2015_CVPR} as the basis, in which the bottleneck layer and the metric network are removed.
The feature networks of multiple views share the same parameters of corresponding convolutional layers. 
The channel number of Conv6 is set to be the same as that of feature maps output by a feature network.
The ReLU activation \cite{2015arXiv150500853X} follows each convolutional layer except the last Conv7.
A layer of L2 normalization is appended after Conv7 whose channel number can be set flexibly to adjust the dimension of descriptors.
The parameters of the full network are detailed in the supplemental material.

\smallskip\noindent\textbf{Loss.}
The two-tower Siamese architecture \cite{simo2015discriminative,chopra2005learning} is adopted here for training. The formulation of the double-margin contrastive loss is used \cite{lin2017deephash}, \ie, 
\iffalse
\begin{equation}
\begin{split}
L(\mathbf{x}_a^{\{1,2,3 \}}, \mathbf{x}_b^{\{1,2,3\}}) & =y \max(||\mathbf{d}_a - \mathbf{d}_b||_2 - \tau_1, 0) \\
& + (1-y) \max(\tau_2 - ||\mathbf{d}_a - \mathbf{d}_b||_2, 0),
\end{split}
\end{equation} 
\fi
\begin{equation}
L(\mathbf{x}_a, \mathbf{x}_b)  =y \max(||\mathbf{d}_a - \mathbf{d}_b||_2 - \tau_1, 0)  + (1-y) \max(\tau_2 - ||\mathbf{d}_a - \mathbf{d}_b||_2, 0),
\end{equation} 
where $y=1$ for positive pairs and $0$ otherwise. $\mathbf{d}_a$ and $\mathbf{d}_b$ are L2-normalized MVDesc descriptors of the two sets of multi-view patches $\mathbf{x}_a$ and $\mathbf{x}_b$, output by the two towers. We set the margins $\tau_1 = 0.3, \tau_2 = 0.6$ in experiments.

\smallskip\noindent\textbf{View number.} Unlike \cite{Su_2015_ICCV,wangdominant,qi2016volumetric} using 12 or 20 views for objects, 
we adopt only 3 views in our MVDesc network for the description of local keypoints, which is a good tradeoff between accuracy and efficiency as shown in Figure \ref{fig:loss}(b).

\smallskip\noindent\textbf{Data preparation.}
Current available patch datasets generally lack sufficient multi-view patches for training. For example, one of the largest training sets Brown \cite{snavely2008modeling,goesele2007multi} only possesses less than 25k 3D points with at least 6 views.
Therefore, we prepare the training data similar to \cite{yi2016lift} based on the self-collected Structure-from-Motion (SfM) database. The database consists of 31 outdoor scenes of urban and rural areas captured by UAV and well reconstructed by a standard 3D reconstruction pipeline \cite{Zhu_2018_CVPR,zhang2018distributed,Zhou_2017_ICCV,shen2016graph,zhu2017parallel,Zhang_2017_ICCV,li2016efficient}.
Each scene contains averagely about 900 images and 250k tracks with at least 6 projections.
The multi-view patches of size 64$\times$64 are cropped from the projections of each track according to SIFT scales and orientations \cite{yi2016lift}, as displayed in Figure \ref{fig:loss}(c).
A positive training pair is formed by two independent sets of triple-view patches from the same track, while a negative pair from different tracks.
A total number of 10 million pairs with equal ratio of positives and negatives are evenly sampled from all the 31 scenes.
We turn the patches into grayscale, subtract the intensities by 128 and divide them by 160 \cite{Han_2015_CVPR}.

\smallskip\noindent\textbf{Training.} We train the network from scratch using SGD with a momentum of 0.9, a weight decay of 0.0005 and a batch size of 256. The learning rate drops by 30\% after every epoch with a base of 0.0001 using exponential decay. The training is generally done within 10 epochs.

\section{Robust Matching Using Belief Propagation (RMBP)} \label{sec:RMBP}
In this section, we are devoted to enhancing the accuracy and robustness of 3D point matching. 
Firstly, a graphical model is defined to describe the spatial organizations of point matches. 
Secondly, each match pair is verified by the inference from the graphical model via belief propagation.
Notably, the proposed method is complementary to the existing matching algorithms \cite{lowe2004distinctive,gold1995new,zhou2016fast,raguram2008comparative}.

\begin{figure}[t]
\begin{center}
\includegraphics[width=1\linewidth]{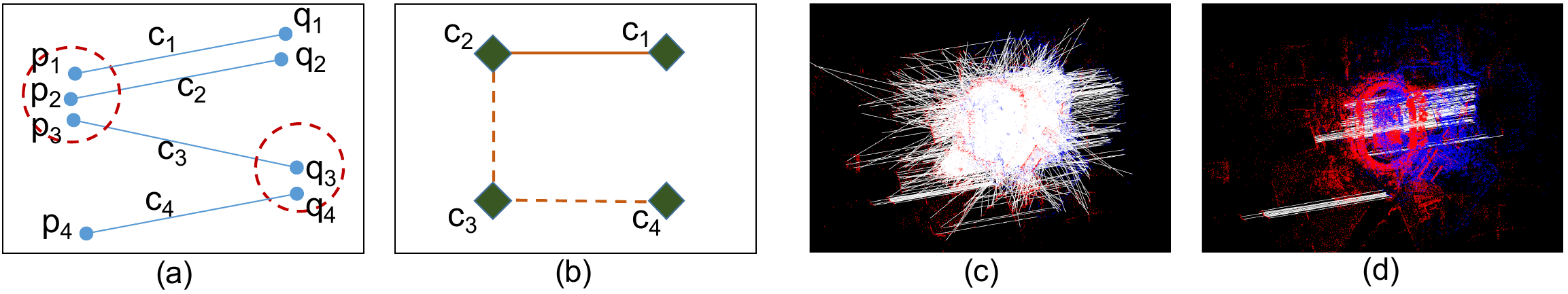}
\end{center}
   \caption{
(a) $c_{1\sim4}$ are four pairs of point matches.
(b) The graph is defined to model the neighboring relationship between $c_{1\sim4}$.
The solid/dashed lines link between compatible/incompatible neighbors.
(c) 4,721 pairs of putative point matches. (d) 109 true match pairs refined by our RMBP}
\label{fig:global_match}
\end{figure}

\subsection{Matching Model}
It can be readily observed that inlier point correspondences generally hold \textit{spatial proximity}.
We illustrate it in Figure \ref{fig:global_match}(a) where $c_1=(\mathbf{p}_1, \mathbf{q}_1)$, $c_2=(\mathbf{p}_2, \mathbf{q}_2)$ and $c_4=(\mathbf{p}_4, \mathbf{q}_4)$ are three pairs of inlier correspondences. 
For any two pairs of inlier matches,
their points in each point cloud are either spatially close to each other like $\langle \mathbf{p}_1, \mathbf{p}_2 \rangle$ and $\langle \mathbf{q}_1, \mathbf{q}_2 \rangle$ or far away from each other like $\langle \mathbf{p}_2, \mathbf{p}_4 \rangle$ and $\langle \mathbf{q}_2, \mathbf{q}_4 \rangle$ at the same time.
On the contrary, outlier correspondences tend to show spatial disorders. This observation implies the probabilistic dependence between neighboring point correspondences which can be modeled by a probabilistic graph.

Formally, we first define the neighborhood of point correspondences as follows. 
Two pairs of point correspondences $c_i = (\mathbf{p}_i, \mathbf{q}_i)$ and $c_j = (\mathbf{p}_j, \mathbf{q}_j)$ are considered as neighbors if either $\mathbf{p}_i$ and $\mathbf{p}_j$, or  $\mathbf{q}_i$ and $\mathbf{q}_j$, are mutually k-nearest neighbors, \ie, 
\begin{align}
\max(\operatorname{rank}(\mathbf{p}_i, \mathbf{p}_j), \operatorname{rank}(\mathbf{p}_j, \mathbf{p}_i)) &< k, \label{eq:match1}\\
\text{or} \,\,\,\, \max(\operatorname{rank}(\mathbf{q}_i, \mathbf{q}_j), \operatorname{rank}(\mathbf{q}_j, \mathbf{q}_i)) &< k, \label{eq:match2}
\end{align}
where $\operatorname{rank}(\mathbf{x}, \mathbf{y})$ denotes the rank of distance of point $\mathbf{y}$ with respect to point $\mathbf{x}$.
Then, the neighboring relationship between $c_i$ and $c_j$ can be further divided into two categories:
if Condition \ref{eq:match1} and \ref{eq:match2} are satisfied simultaneously, $c_i$ and $c_j$ are called \textit{compatible} neighbors. They are very likely to co-exist as inlier matches. 
But if only one of Condition \ref{eq:match1} or \ref{eq:match2} is satisfied by one point pair but another pair of points in the other point cloud locate far apart from each other, \eg, 
\begin{align}
\min(\operatorname{rank}(\mathbf{p}_i, \mathbf{p}_j), \operatorname{rank}(\mathbf{p}_j, \mathbf{p}_i)) &> l, 
\label{eq:match3}\\
\text{or} \,\,\,\, \min(\operatorname{rank}(\mathbf{q}_i, \mathbf{q}_j), \operatorname{rank}(\mathbf{q}_j, \mathbf{q}_i)) &> l,
\label{eq:match4}
\end{align}
$c_i$ and $c_j$ are called \textit{incompatible} neighbors, as it is impossible for two match pairs breaching spatial proximity to be inliers simultaneously. 
The threshold parameter $k$ in Condition \ref{eq:match1} and \ref{eq:match2} is set to a relatively small value,
while the parameter $l$ in Condition \ref{eq:match3} and \ref{eq:match4} is set to be larger than $k$ by a considerable margin. 
These settings are intended to ensure sufficiently strict conditions on identifying compatible or incompatible neighbors for robustness.

Based on the spatial property of point matches stated above, an underlying graphical model is built to model the pairwise interactions between neighboring match pairs, as shown in Figure \ref{fig:global_match}(a) and (b). 
The nodes in graphical model are first defined as a set of binary variables $\mathcal{X} = \{x_i \}$ each associated with a pair of point correspondence. $x_i \in \{0, 1 \}$ indicates the latent state of being an outlier or inlier, respectively.
Then the undirected edges between nodes are formed based on the compatible and incompatible neighboring relationship defined above.
With the purpose of rejecting outliers, the objective here is to compute the marginal of being an inlier for each point correspondence by performing inference on the defined model.

\subsection{Inference by Belief Propagation}
The task of computing marginals on nodes of a cyclic network is known to be NP-hard \cite{cooper1990computational}. As a disciplined inference algorithm, loopy belief propagation (LBP) provides approximate yet compelling inference on arbitrary networks \cite{murphy1999loopy}. 

In the case of our graphical network with binary variables and pairwise interactions, the probabilistic distributions of all node variables are first initialized as $[0.5, 0.5]^T$ with no prior imposed.
Then the iterative message update step of a standard LBP algorithm at iteration $t$ can be written as 
\begin{equation} \label{eq:message}
\mathbf{m}_{ij}^{t+1} = \frac{1}{Z} \mathbf{F}_{ij} \mathbf{m}_i \prod_{k \in \partial i \backslash j} \mathbf{m}_{ki}^t.
\end{equation}
Here, $\partial i$ denotes the set of neighbors of node $x_i$ and $Z$ is the L1 norm of the incoming message for normalization. The message $\mathbf{m}_{ij}$ passed from node $x_i$ to $x_j$ is a two-dimensional vector, which represents the belief of $x_j$'s probability distribution inferred by $x_i$. So is the constant message $\mathbf{m}_i$ passed from the observation of node $x_i$, which indicates the likelihood distribution of $x_i$ given its observation measurements like descriptor similarity. 
The first and second components of the messages are the probabilities of being an outlier and an inlier, respectively.
The product of messages is component-wise. The $2 \times 2$ matrix $\mathbf{F}_{ij}$ is the compatibility matrix of node $x_i$ and $x_j$. Based on the neighboring relationship analyzed above, the compatibility matrix is supposed to favor the possibility that both nodes are inliers if they are compatible neighbors and the reverse if they are incompatible neighbors. In order to explicitly specify the bias, the compatibility matrices take two forms in the two different cases, respectively:
\begin{equation}\label{eq:compatibility}
\mathbf{F}^+=
\begin{bmatrix}
    1 & 1 \\
    1 & \lambda
\end{bmatrix}
\quad\text{or}\quad
\mathbf{F}^-=
\begin{bmatrix}
    \lambda & \lambda \\
    \lambda & 1
\end{bmatrix} .
\end{equation}
The parameter $\lambda$ takes a biased value greater than $1$. To guarantee the convergence of LBP, Simon's condition \cite{tatikonda2002loopy} is enforced and the value of $\lambda$ is thus constrained by 
\begin{equation} \label{eq:condition}
\max_{x_i \in \mathcal{X}} |\partial i| \cdot \log\lambda < 2,
\end{equation}
in which way $\lambda$ is set adaptively according to the boundary condition.
The proof of the convergence's condition is detailed in the supplemental material.
After convergence, the marginal distribution of node $x_i$ is derived by
\begin{equation} 
\mathbf{b}_i = \frac{1}{Z} \mathbf{m}_i \prod_{k \in \partial i} \mathbf{m}_{ki},
\end{equation}
which unifies implication from individual observations and beliefs from structured local neighborhood. 
After the inference, point matches with low marginals (\eg $<0.5$) are discarded as outliers. It greatly contributes to the matching accuracy as shown in Figure \ref{fig:global_match}(d) where 109 true match pairs are refined from 4,721 noisy putative match pairs.

\smallskip\noindent\textbf{Complexity analysis.}
The complexity of LBP is known to be linear to the number of edges in the graph \cite{yedidia2003understanding}. And the Condition \ref{eq:match1} and \ref{eq:match2} bound the degree of each node to be less than $2k$, so that the upper bound of RMBP's complexity is linear with the number of nodes.

\section{Experiments}
In this section, we first individually evaluate the proposed MVDesc and RMBP in Section \ref{sec:eval_mvdesc} and \ref{sec:eval_rmbp} respectively. Then the two approaches are validated on the tasks of geometric registration in Section \ref{sec:eval_reg}.

All the experiments, including the training and testing of neural networks, are done on a machine with a 8-core Intel i7-4770K, a 32GB memory and a NVIDIA GTX980 graphics card. In the experiments below, when we say putative matching, we mean finding the correspondences of points whose descriptors are mutually closest to each other in Euclidean space between two point sets. The matching is implemented based on OpenBLAS \cite{openblas}. 
The traditional geometric descriptors \cite{rusu2008aligning,rusu2009fast,tombari2010SHOT,tombari2010USC}
are produced based on PCL \cite{Rusu_ICRA2011_PCL}.

\subsection{Evaluation of MVDesc} 
\label{sec:eval_mvdesc}
The target here is to compare the description ability of the proposed MVDesc against the state-of-the-art patch descriptors \cite{lowe2004distinctive,simo2015discriminative,tian2017l2} and geometric descriptors \cite{rusu2008aligning,rusu2009fast,tombari2010SHOT,tombari2010USC,zeng20163dmatch,Khoury_2017_ICCV}.

\smallskip\noindent\textbf{5.1.1 \space\space Comparisons with patch descriptors}

\smallskip\noindent\textbf{Setup.}
We choose HPatches \cite{hpatches_2017_cvpr}, one of the largest local patch benchmarks, for evaluation. It consists of 59 cases and 96,315 6-view patch sets.
First, we partition each patch set into two subsets by splitting the 6-views into halves. 
Then, the 3-view patches are taken as input to generate descriptors.
We set up the three benchmark tasks in \cite{hpatches_2017_cvpr} by reorganizing the 3-view patches and use the mean average precision (mAP) as measurement.
For the patch verification task, we collect all the 96,315 positive pairs of 3-view patches and 100,000 random negatives.
For the image matching task, we apply putative matching across the two half sets of each case after mixing 6,000 unrelated distractors into every half. 
For the patch retrieval task, we use 
the full 96,315 6-view patch sets each of which corresponds to a 3-view patch set as a query and the other 3-view set in the database. Besides, we mix 100,000 3-view patch sets from an independent image set into the database for distraction.

We make comparisons with the baseline SIFT \cite{lowe2004distinctive} and the state-of-the-art DeepDesc \cite{simo2015discriminative} and L2-Net \cite{tian2017l2}, for which we randomly choose a single view from the 3-view patches.
To verify the advantage of our FRN over the widely-used view pooling \cite{Su_2015_ICCV,wangdominant,qi2016volumetric} in multi-view fusion, we remove the Fuseption branch from our MVDesc network and train with the same data and configuration.
All the descriptors have the dimensionality of 128.

\begin{table}[]
\centering
\caption{The mAPs of descriptors in the three tasks of HPatches benchmark \cite{hpatches_2017_cvpr}. 
Our MVDesc holds the top place in all the three tasks}
\label{table:hpatch}
\resizebox{0.8\linewidth}{!}
{
\begin{tabular}{|c|c|c|c|c|c|c|}
\hline
      & SIFT \cite{lowe2004distinctive}  & DeepDesc \cite{simo2015discriminative} & L2-Net \cite{tian2017l2} & View pooling \cite{Su_2015_ICCV}  & MVDesc     \\ \hline
Patch verification  & 0.646 & 0.716    & 0.792  & 0.883 & \textbf{0.921}   \\ \hline
Image matching  & 0.111 & 0.172    & 0.309  & 0.312 & \textbf{0.325}   \\ \hline
Patch retrieval  & 0.269 & 0.357    & 0.414  & 0.456 & \textbf{0.530}   \\ \hline
\end{tabular}
}
\end{table}

\begin{figure}[t]
\begin{center}
\includegraphics[width=0.8\linewidth]{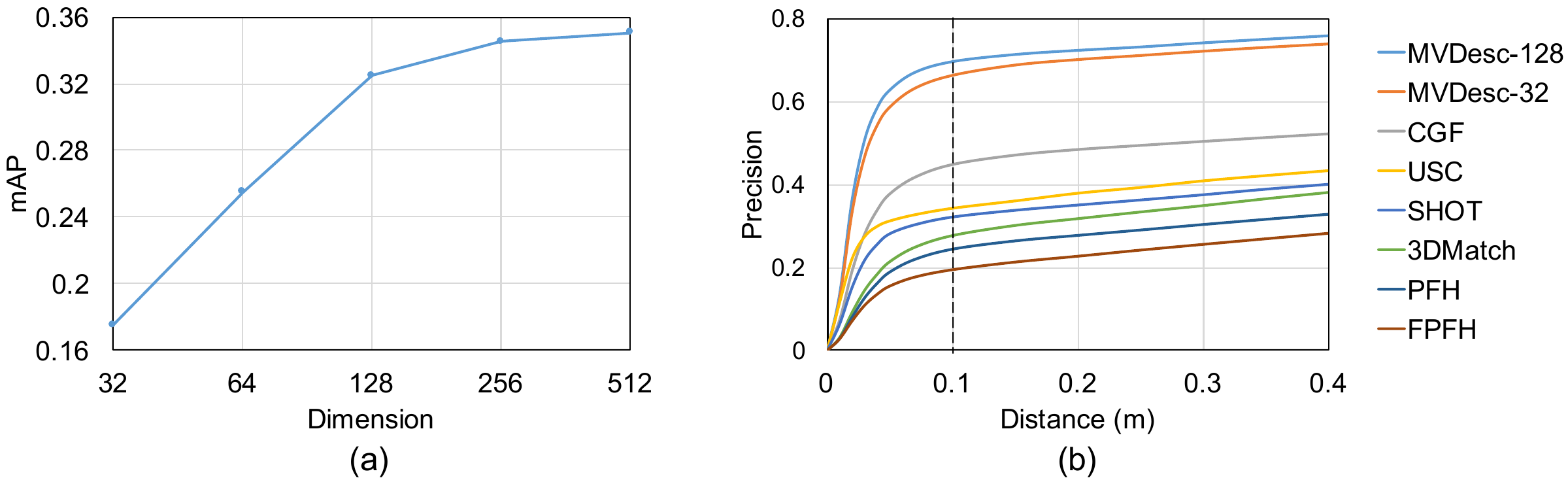}
\end{center}
   \caption{(a) The trade-off of mAP versus dimension of our MVDesc on the HPatches benchmark \cite{hpatches_2017_cvpr}; (b) The change of matching precisions \wrt the varying threshold of point distances on the TUM dataset \cite{sturm2012benchmark}. The 128- and 32-dimensional MVDesc rank first and second at any threshold}
\label{fig:geo_desc}
\end{figure}

\smallskip\noindent\textbf{Results.}
The statistics in Table \ref{table:hpatch} show that our MVDesc achieves the highest mAPs in all the three tasks. 
First, it demonstrates the advantage of our FRN over view pooling \cite{Su_2015_ICCV,wangdominant,qi2016volumetric,huang2018learning,chen2016multi} in terms of multi-view fusion. 
Second, the improvement of MVDesc over DeepDesc \cite{simo2015discriminative} and L2-Net \cite{tian2017l2} suggests the benefits of leveraging more image views than a single one.
Additionally, we illustrate in Figure \ref{fig:geo_desc}(a) the trade-off between the mAP of the image matching task and the dimension of our MVDesc. The mAP rises but gradually saturates with the increase of dimension.

\smallskip\noindent\textbf{5.1.2 \space\space Comparisons with geometric descriptors} 

\smallskip\noindent\textbf{Setup.}
Here, we perform evaluations on matching tasks of the RGB-D dataset TUM \cite{sturm2012benchmark}.
Following \cite{Khoury_2017_ICCV}, we collect up to 3,000 pairs of overlapping point cloud fragments from 10 scenes of TUM. Each fragment is recovered from independent RGB-D sequences of 50 frames. 
We detect keypoints from the fragments by SIFT3D \cite{flint2007thrift} and then generate geometric descriptors, including PFH \cite{rusu2008aligning}, FPFH \cite{rusu2009fast}, SHOT \cite{tombari2010SHOT}, USC \cite{tombari2010USC}, 3DMatch \cite{zeng20163dmatch} and CGF \cite{Khoury_2017_ICCV}. Our MVDesc is derived from the projected patches of keypoints in three randomly-selected camera views.
For easier comparison, two dimensions, 32 and 128, of MVDesc (MVDesc-32 and MVDesc-128) are adopted.
Putative matching is applied to all the fragment pairs to obtain correspondences.
Following \cite{Khoury_2017_ICCV}, we measure the performance of descriptors by the fraction of correspondences whose ground-truth distances lie below the given threshold.

\smallskip\noindent\textbf{Results.}
The precision of point matches \wrt the threshold of point distances is depicted in Figure \ref{fig:geo_desc}(b). The MVDesc-128 and MVDesc-32 rank first and second in precision at any threshold, outperforming the state-of-the-art works by a considerable margin. 
We report in Table \ref{table:geo_desc} the precisions and recalls when setting the threshold to 0.1 meters and the average time of producing 1,000 descriptors.
Producing geometric descriptors in general is slower than MVDesc due to the cost of computing local histograms, although the computation has been accelerated by multi-thread parallelism.

\begin{table}[]
\centering
\caption{Precisions and recalls of matching on the TUM dataset \cite{sturm2012benchmark} when the threshold of points' distances equals to 0.1 meters. The average time taken to encode 1,000 descriptors is also compared. Our MVDesc hits the best in terms of precision, recall and efficiency}
\label{table:geo_desc}
\resizebox{0.9\linewidth}{!}
{
\begin{tabular}{|c|c|c|c|c|c|c|c|c|}
\hline
          & CGF \cite{Khoury_2017_ICCV}   & 
          FPFH \cite{rusu2009fast} & 
          PFH \cite{rusu2008aligning}  & 
          SHOT \cite{tombari2010SHOT} & 
          3DMatch \cite{zeng20163dmatch} & 
          USC \cite{tombari2010USC}   & 
          \multicolumn{2}{c|}{MVDesc}    \\ \hline
Dim.      & 32    & 33    & 125   & 352   & 512     & 1980  & 32            & 128            \\ \hline
Precision & 0.447 & 0.194 & 0.244 & 0.322 & 0.278   & 0.342 & 0.664         & \textbf{0.695} \\ \hline
Recall & 0.215 & 0.229 & 0.265 & 0.093 & 0.114   & 0.026 & 0.523         & \textbf{0.580} \\ \hline
Time (s)  & 7.60  & 1.49  & 14.40  & 0.29  & 2.60     & 0.73  & \textbf{0.22} & 0.23           \\ \hline
\end{tabular}
}
\end{table}

\subsection{Evaluation of RMBP}  \label{sec:eval_rmbp}

\begin{figure}[t]
\begin{center}
\includegraphics[width=0.85\linewidth]{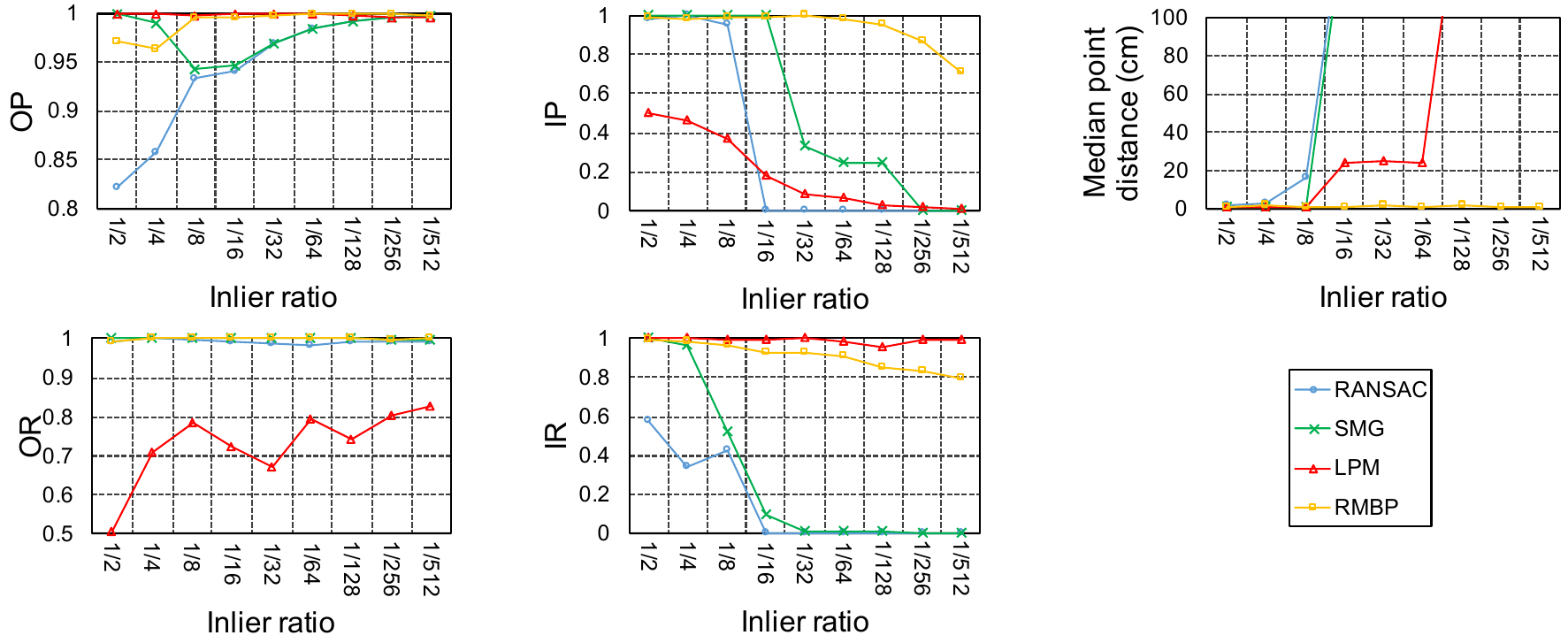}
\end{center}
   \caption{The mean precisions and recalls of outlier rejection (OP, OR) and inlier selcection (IP, IR), as well as median point distance after registration, \wrt the inlier ratio. RMBP performs well in all metrics and at all inlier ratios while RANSAC, SMG \cite{rodola2013scale} and LPM \cite{ijcai2017-627} fail to give valid registrations when the inlier ratio drops below $\frac{1}{8}$, $\frac{1}{8}$ and $\frac{1}{64}$, respectively}
\label{fig:rmbp_cmp}
\end{figure}

\smallskip\noindent\textbf{Setup.}
To evaluate the performance of outlier rejection, we compare RMBP with RANSAC and two state-of-the-art works - Sparse Matching Game (SMG) \cite{rodola2013scale} and Locality Preserving Matching (LPM) \cite{ijcai2017-627}.  
All the parameters of methods have been tuned to give the best results.
We match 100 pairs of same-scale point clouds from 20 diverse indoor and outdoor scenes of TUM [27], ScanNet [65] and EPFL [38] datasets. 
We keep a constant number of inlier correspondences and continuously add outlier correspondences for distraction.
The evaluation uses the metrics: 
the mean precisions and recalls of outlier rejection and inlier selection.
Formally, we write 
$OP=\frac{\#true\,\, rejections}{\#rejections}$, $OR=\frac{\#true\,\, rejections}{\#outliers}$, 
$IP=\frac{\#kept\,\, inliers}{\#kept\,\, matches}$, $IR=\frac{\#kept\,\, inliers}{\#inliers}$.
We also estimate the transformations from kept matches using RANSAC and collect the median point distance after registration.

\smallskip\noindent\textbf{Results.}
The measurements with respect to the inlier ratio are shown in Fig. \ref{fig:rmbp_cmp}.
First, RMBP is the only method achieving high performance in all metrics and at all inlier ratios.
Second, RANSAC, SMG and LPM fail to give valid registrations when the inlier ratio drops below $\frac{1}{8}$, $\frac{1}{8}$ and $\frac{1}{64}$, respectively.
They obtain high OP and OR but low IP or IR when the inlier ratio is smaller than $\frac{1}{16}$, because they tend to reject almost all the matches.

\iffalse
\begin{table}[h]
\centering
\label{table:match-time}
\begin{tabular}{|c|c|c|c|}
\hline
Stages           & Putative matching & RMBP  & RANSAC \\ \hline
Time (ms) & 221.42         & 1.55 & 0.81   \\ \hline
\end{tabular}
\caption{The average run-time of three stages of matching two point sets each with 5k 128-dimensional descriptors. Our RMBP shows acceptable cost - about 1/140 of that of putative matching}
\label{table:RMBP_efficiency}
\end{table} 
\fi

\subsection{Geometric Registration}  \label{sec:eval_reg}
In this section, we verify the practical usage of the proposed MVDesc and RMBP by the tasks of geometric registration. We operate on point cloud data obtained from two different sources: the point clouds scanned by RGB-D sensors and those reconstructed by multi-view stereo (MVS) algorithms \cite{schoenberger2016mvs}.

\smallskip\noindent\textbf{5.3.1 \space\space Registration of scanning data}

\smallskip\noindent\textbf{Setup.} We use the task of loop closure as in \cite{zeng20163dmatch,Khoury_2017_ICCV} based on the dataset ScanNet \cite{dai2017scannet}, where we check whether two overlapping sub-maps of an indoor scene can be effectively detected and registered.
Similar to \cite{zeng20163dmatch,Khoury_2017_ICCV}, we build up independent fragments of 50 sequential RGB-D frames from 6 different indoor scenes of ScanNet \cite{dai2017scannet}. For each scene, we collect more than 500 fragment pairs with labeled overlap obtained from the ground truth for registration. 

The commonly-used registration algorithm, putative matching plus RANSAC, is adopted in combination with various descriptors \cite{rusu2008aligning,rusu2009fast,tombari2010SHOT,tombari2010USC,zeng20163dmatch,Khoury_2017_ICCV}. The proposed RMBP serves as an optional step before RANSAC. We use the same metric as \cite{zeng20163dmatch,Khoury_2017_ICCV}, \ie, the precision and recall of registration of fragment pairs. Following \cite{zeng20163dmatch}, a registration is viewed as true positive if the estimated Euclidean transformation yields more than $30\%$ overlap between registered fragments and transformation error less then $0.2m$. We see a registration as positive if there exist more than 20 pairs of point correspondences after RANSAC.

\begin{figure}[t]
\begin{center}
\includegraphics[width=1.0\linewidth]{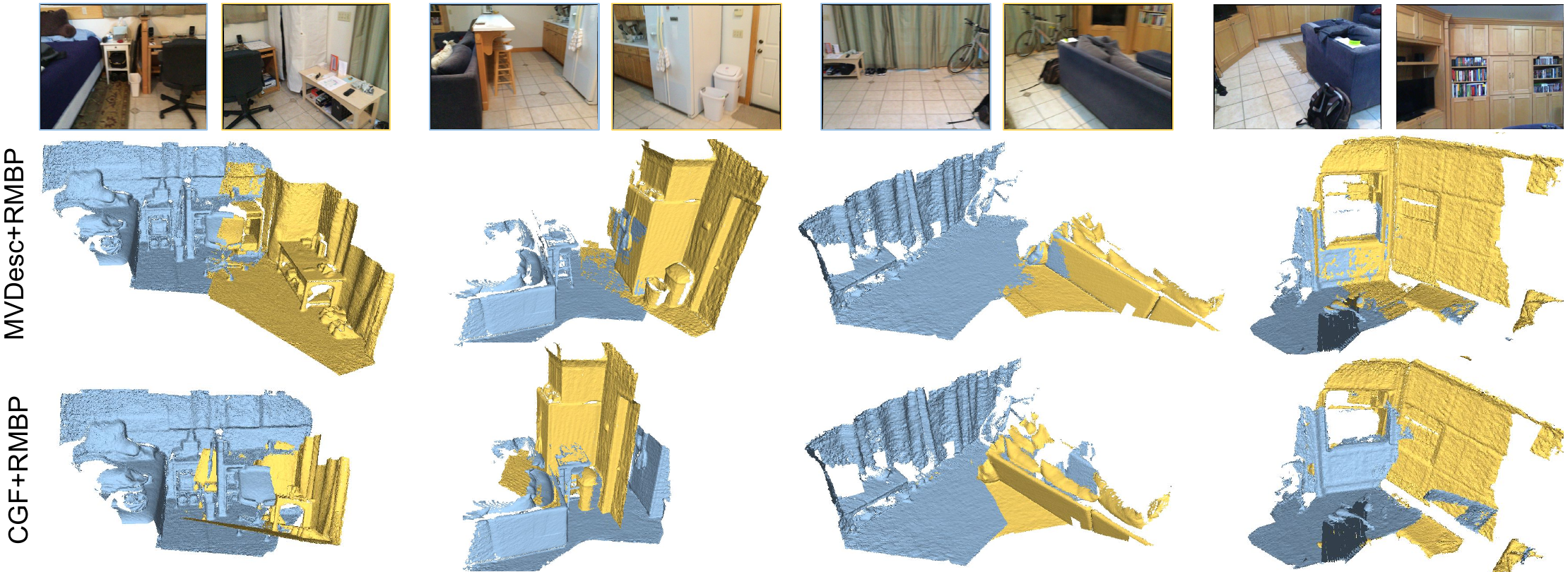}
\end{center}
   \caption{Challenging cases of loop closures from the ScanNet dataset \cite{dai2017scannet}. The images in the top row indicate very limited overlap shared by the fragment pairs. Our MVDesc-32+RMBP succeeds in the registration of these cases while the top-performing geometric descriptor CGF-32 \cite{Khoury_2017_ICCV} fails no matter whether RMBP is employed}
\label{fig:fragments}
\end{figure}

\smallskip\noindent\textbf{Results.} The precisions, recalls and the average time of registration per pair are reported in Table \ref{table:geo_registration}. 
Our MVDesc-32 and MVDesc-128 both surpass the counterparts by a significant margin in recall while with comparable precision and efficiency. Our versatile RMBP well improves the precisions for 6 out of 8 descriptors and lifts the recalls for 7 out of 8 descriptors. 
The sample registration results of overlap-deficient fragments are visualized in Figure \ref{fig:fragments}.

\begin{table}[]
\centering
\caption{The quantitative comparisons of 3D descriptors on registration of the ScanNet dataset \cite{dai2017scannet}. The superscript * means the proposed RMBP is applied. The RMBP generally lifts the precisions and recalls of registration for almost all the descriptors. Our MVDesc well surpasses the state-of-the-art works in recall with comparable precision and run-time of registration}
\label{table:geo_registration}
\resizebox{0.9\linewidth}{!}
{
\begin{tabular}{|c|c|c|c|c|c|c|c|c|}
\hline          
           & CGF \cite{Khoury_2017_ICCV}
           & FPFH \cite{rusu2009fast}
           & PFH \cite{rusu2008aligning}  
           & SHOT \cite{tombari2010SHOT}
           & 3DMatch \cite{zeng20163dmatch}
           & USC \cite{tombari2010USC} 
           & \multicolumn{2}{c|}{MVDesc} \\ \hline
Dim.       & 32             & 33    & 125   & 352   & 512            & 1980  & 32      & 128             \\ \hline
Precision  & 0.914 & 0.825 & 0.866 & 0.875 & 0.890          & 0.790 & 0.865   & 0.910           \\ \hline
Precision* & 0.927          & 0.856 & 0.864 & 0.928 & 0.934 & 0.795 & 0.892   & 0.906           \\ \hline
Recall     & 0.350          & 0.119 & 0.147 & 0.178 & 0.185          & 0.124 & 0.421   & 0.490 \\ \hline
Recall*    & 0.419          & 0.272 & 0.338 & 0.198 & 0.145          & 0.157 & 0.482   & 0.513  \\ \hline
Time (s)    & 0.5            & 0.5   & 0.7   & 1.8   & 2.4            & 8.4   & 0.5     & 0.7             \\ \hline
\end{tabular}
}
\end{table}

\begin{figure}[t]
\begin{center}
\includegraphics[width=1.0\linewidth]{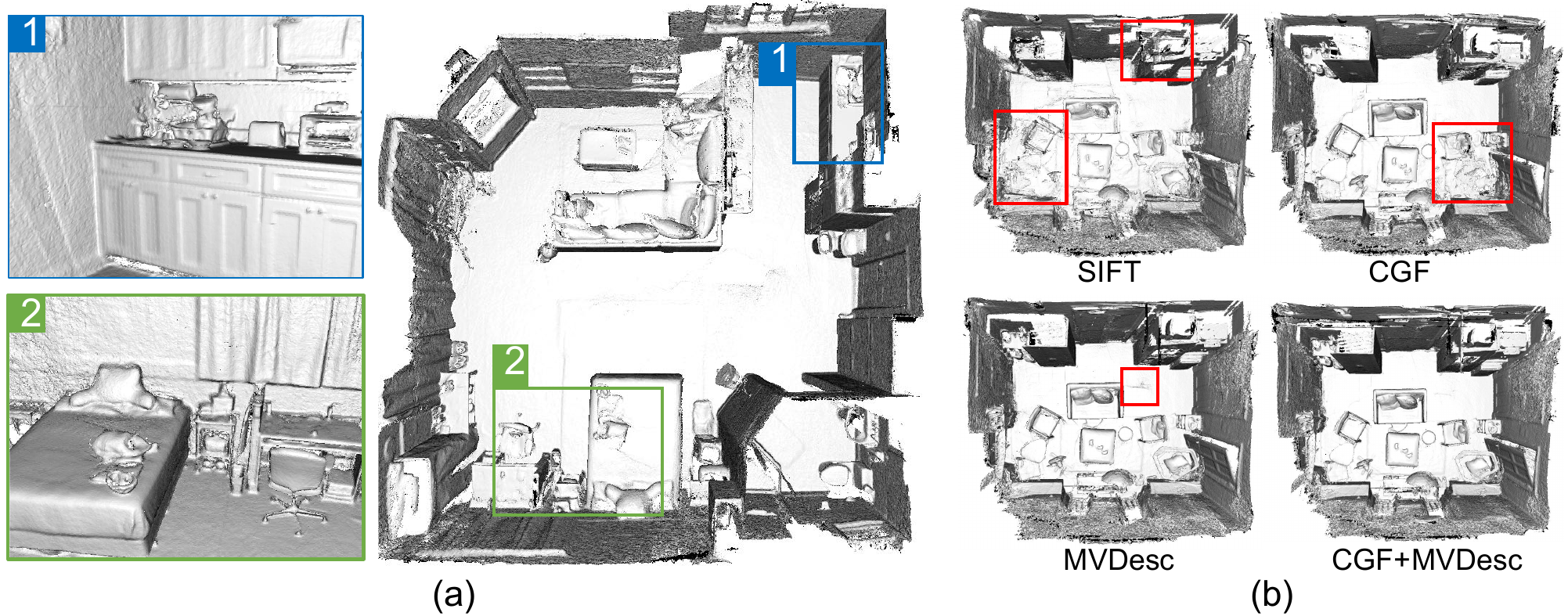}
\end{center}
   \caption{(a) A complete reconstruction of an apartment from ScanNet \cite{dai2017scannet} using our MVDesc-32. (b) The reconstructions using SIFT \cite{lowe2004distinctive}, CGF-32 \cite{Khoury_2017_ICCV} and MVDesc-32. The collaboration of MVDesc-32 and CGF-32 yields the best reconstruction as shown in the last cell of (b)}
\label{fig:indoor}
\end{figure}

\smallskip\noindent\textbf{Indoor reconstruction.} The practical usage of MVDesc is additionally evaluated by indoor reconstruction of the ScanNet dataset \cite{dai2017scannet}.
We first build up reliable local fragments through RGB-D odometry following \cite{kerl2013robust,Choi_2015_CVPR} and then globally register the fragments based on \cite{zhou2016fast}. 
The RMBP is applied for outlier filtering.
The FPFH \cite{rusu2009fast} used in \cite{zhou2016fast} is replaced by SIFT \cite{lowe2004distinctive}, CGF-32 \cite{Khoury_2017_ICCV} or MVDesc-32 to establish correspondences. We also test the collaboration of CGF-32 and MVDesc-32 by combining their correspondences. 
Our MVDesc-32 contributes to visually compelling reconstruction as shown in Figure \ref{fig:indoor}(a). And we find that MVDesc-32 functions as a solid complement to CGF-32 as shown in Figure \ref{fig:indoor}(b), especially for the scenarios with rich textures.

\smallskip\noindent\textbf{5.3.2 \space\space Registration of Multi-View Stereo (MVS) data}

\begin{figure}[t]
\begin{center}
\includegraphics[width=0.85\linewidth]{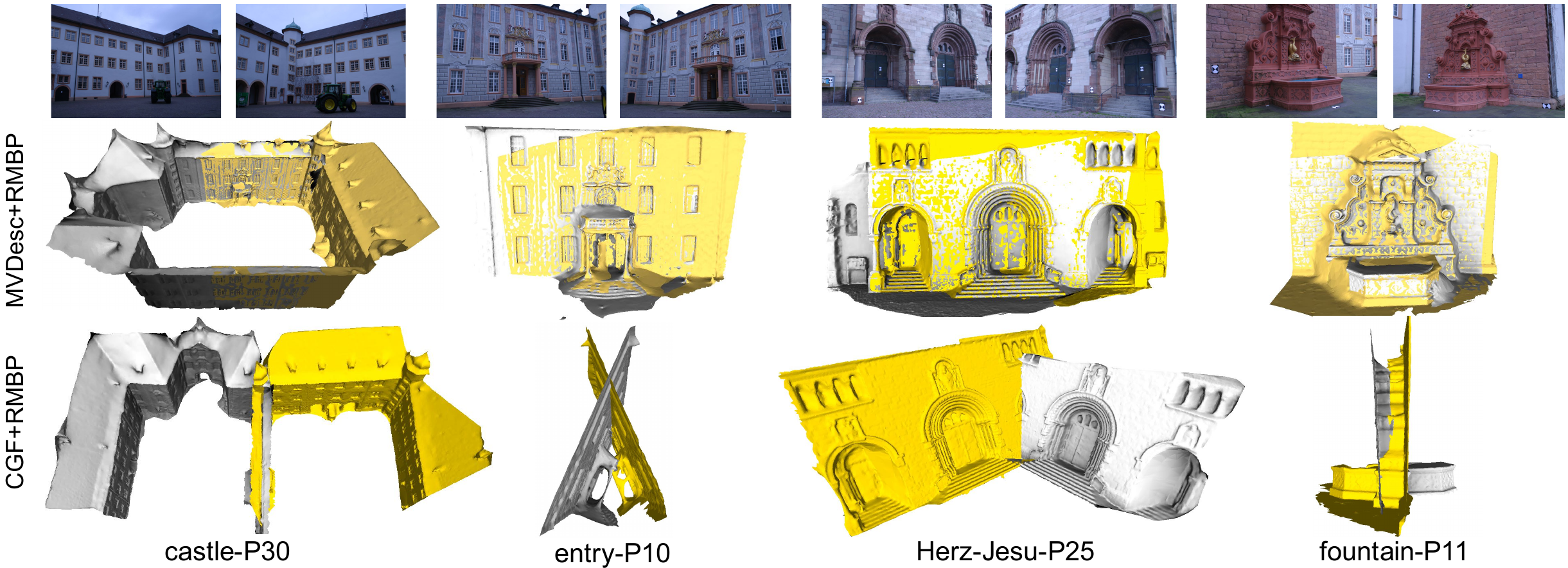}
\end{center}
   \caption{Registrations of models of EPFL benchmark \cite{strecha2008benchmarking}. Given the same keypoints, our MVDesc-32+RMBP accomplishes correct registrations while CGF-32 \cite{Khoury_2017_ICCV}+RMBP fails due to the symmetric ambiguity of the geometry}
\label{fig:epfl_reg}
\end{figure}

\smallskip\noindent\textbf{Setup.}
Aside from the scanning data, we run registration on the four scenes of the MVS benchmark EPFL \cite{strecha2008benchmarking}.
First, we split the cameras of each scene into two clusters in space highlighting the difference between camera views, as shown in Figure \ref{fig:epfl_reg}. 
Then, the ground-truth camera poses of each cluster are utilized to independently reconstruct the dense point clouds by the MVS algorithm \cite{schoenberger2016mvs}.
Next, we detect keypoints by SIFT3D \cite{flint2007thrift} and generate descriptors \cite{rusu2008aligning,rusu2009fast,tombari2010SHOT,tombari2010USC,zeng20163dmatch,Khoury_2017_ICCV} for each point cloud.
The triple-view patches required by MVDesc-32 are obtained by projecting keypoints into 3 visible image views randomly with occlusion test by ray racing \cite{wald2001interactive}.
After, the correspondences between the two point clouds of each scene are obtained by putative matching and then RMBP filtering.
Finally, we register the two point clouds of each scene based on FGR \cite{zhou2016fast} using estimated correspondences.

\begin{figure}[t]
\begin{center}
\includegraphics[width=0.85\linewidth]{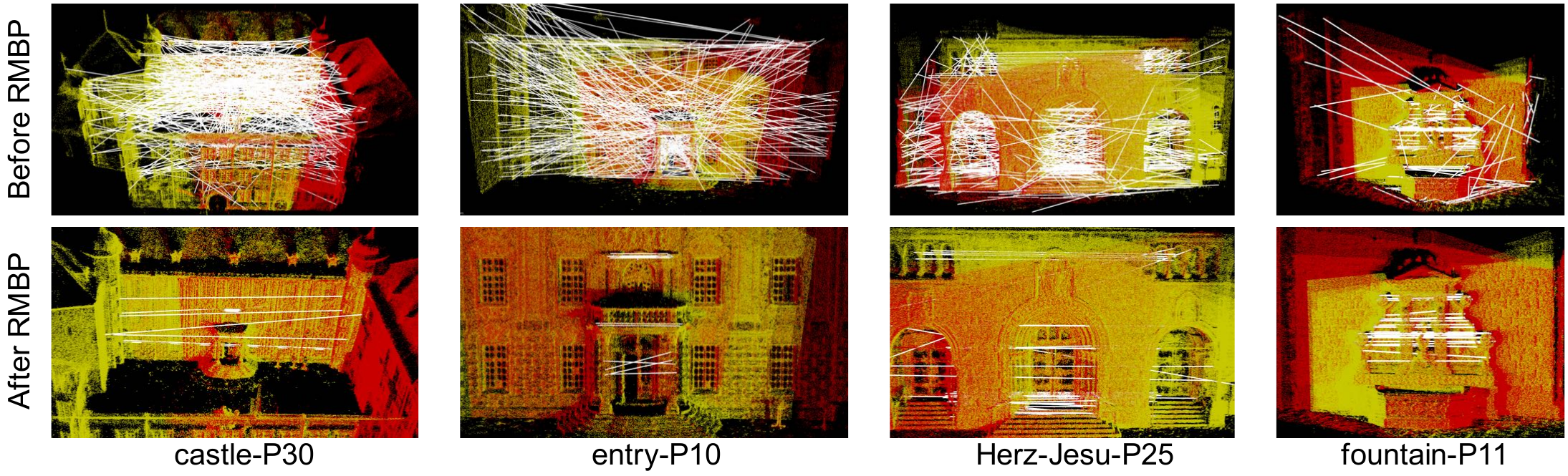}
\end{center}
   \caption{
Spurious correspondences of CGF features \cite{Khoury_2017_ICCV} before and after RMBP filtering. The two point clouds of \cite{strecha2008benchmarking} (colored in yellow and red) have been overlaid together by the ground truth transformation for visualization.
Although our RMBP has eliminated most of the unorganized false matches, it is incapable of rejecting those ambiguous outliers arising from the ambiguity of the symmetric geometry}
\label{fig:epfl_match}
\end{figure}

\smallskip\noindent\textbf{Results.}
Our MVDesc-32 and RMBP help to achieve valid registrations for all the four scenes, whilst none of the geometric descriptors including CGF \cite{Khoury_2017_ICCV}, 3DMatch \cite{zeng20163dmatch}, FPFH \cite{rusu2009fast}, PFH \cite{rusu2008aligning}, SHOT \cite{tombari2010SHOT} and USC \cite{tombari2010USC} do, as shown in Figure \ref{fig:epfl_reg}. 
It is found that the failure is mainly caused by the geometrically symmetric patterns of the four scenes. 
We show the correspondences between CGF-32 features \cite{Khoury_2017_ICCV} in Figure \ref{fig:epfl_match} as an example. The putative matching has resulted in a large number of ambiguous correspondences between keypoints located at the symmetric positions. 
And in essence, our RMBP is incapable of resolving the ambiguity in such cases though, because the correspondences in a symmetric structure still adhere to the geometric consistency. Ultimately, the ambiguous matches lead to the horizontally-flipped registration results as shown in Figure \ref{fig:epfl_reg}. 
At least in the EPFL benchmark \cite{strecha2008benchmarking}, the proposed MVDesc descriptor shows superior ability of description to the geometric ones.

\section{Conclusion}
In this paper, we address the correspondence problem for the registration of point clouds. 
First, a multi-view descriptor, named MVDesc, has been proposed for the encoding of 3D keypoints, which strengthens the representation by applying the fusion of image views \cite{Su_2015_ICCV,wangdominant,qi2016volumetric} to patch descriptor learning \cite{simo2015discriminative,tian2017l2,han2015matchnet,yi2016lift,balntas2016learning}.
Second, a robust matching method, abbreviated as RMBP, has been developed to resolve the rejection of outlier matches by means of efficient inference through belief propagation \cite{murphy1999loopy} on the defined graphical matching model.
Both approaches have been validated to be conductive to forming point correspondences of better quality for registration, as demonstrated by the intensive comparative evaluations and registration experiments \cite{rusu2008aligning,rusu2009fast,tombari2010SHOT,tombari2010USC,zeng20163dmatch,Khoury_2017_ICCV,zhou2016fast,simo2015discriminative,tian2017l2,raguram2008comparative}. 

\smallskip\noindent\textbf{Acknowledgement.}
This work is supported by Hong Kong RGC 16208614, T22-603/15N, Hong Kong ITC PSKL12EG02, and China 973 program, 2012CB-\\316300.

\bibliographystyle{splncs}
\bibliography{egbib}

\begin{thebibliography}{10}

\bibitem{Zhu_2018_CVPR}
Zhu, S., Zhang, R., Zhou, L., Shen, T., Fang, T., Tan, P., Quan, L.:
\newblock Very large-scale global sfm by distributed motion averaging.
\newblock In: CVPR. (2018)

\bibitem{zhang2018distributed}
Zhang, R., Zhu, S., Shen, T., Zhou, L., Luo, Z., Fang, T., Quan, L.:
\newblock Distributed very large scale bundle adjustment by global camera
  consensus.
\newblock PAMI (2018)

\bibitem{dissanayake2001solution}
Dissanayake, M.G., Newman, P., Clark, S., Durrant-Whyte, H.F., Csorba, M.:
\newblock A solution to the simultaneous localization and map building (slam)
  problem.
\newblock IEEE Transactions on Robotics and Automation \textbf{17}(3) (2001)
  229--241

\bibitem{fraundorfer2012visual}
Fraundorfer, F., Scaramuzza, D.:
\newblock Visual odometry: Part ii: Matching, robustness, optimization, and
  applications.
\newblock IEEE Robotics \& Automation Magazine \textbf{19}(2) (2012)  78--90

\bibitem{johnson1999using}
Johnson, A.E., Hebert, M.:
\newblock Using spin images for efficient object recognition in cluttered 3d
  scenes.
\newblock PAMI \textbf{21}(5) (1999)

\bibitem{rusu2008aligning}
Rusu, R.B., Blodow, N., Marton, Z.C., Beetz, M.:
\newblock Aligning point cloud views using persistent feature histograms.
\newblock In: IROS. (2008)

\bibitem{rusu2009fast}
Rusu, R.B., Blodow, N., Beetz, M.:
\newblock Fast point feature histograms (fpfh) for 3d registration.
\newblock In: ICRA. (2009)

\bibitem{tombari2010SHOT}
Tombari, F., Salti, S., Di~Stefano, L.:
\newblock Unique signatures of histograms for local surface description.
\newblock In: ECCV. (2010)

\bibitem{tombari2010USC}
Tombari, F., Salti, S., Di~Stefano, L.:
\newblock Unique shape context for 3d data description.
\newblock In: Proceedings of the ACM workshop on 3D object retrieval. (2010)

\bibitem{zeng20163dmatch}
Zeng, A., Song, S., Nie{\ss}ner, M., Fisher, M., Xiao, J., Funkhouser, T.:
\newblock 3dmatch: Learning local geometric descriptors from rgb-d
  reconstructions.
\newblock In: CVPR. (2017)

\bibitem{Khoury_2017_ICCV}
Khoury, M., Zhou, Q.Y., Koltun, V.:
\newblock Learning compact geometric features.
\newblock In: ICCV. (2017)

\bibitem{deng2018ppfnet}
Deng, H., Birdal, T., Ilic, S.:
\newblock Ppfnet: Global context aware local features for robust 3d point
  matching.
\newblock arXiv preprint (2018)

\bibitem{lowe2004distinctive}
Lowe, D.G.:
\newblock Distinctive image features from scale-invariant keypoints.
\newblock IJCV \textbf{60}(2) (2004)  91--110

\bibitem{gold1995new}
Gold, S., Lu, C.P., Rangarajan, A., Pappu, S., Mjolsness, E.:
\newblock New algorithms for 2d and 3d point matching: Pose estimation and
  correspondence.
\newblock In: Advances in Neural Information Processing Systems. (1995)

\bibitem{zhou2016fast}
Zhou, Q.Y., Park, J., Koltun, V.:
\newblock Fast global registration.
\newblock In: ECCV. (2016)

\bibitem{2018arXiv180207869G}
{Georgakis}, G., {Karanam}, S., {Wu}, Z., {Ernst}, J., {Kosecka}, J.:
\newblock End-to-end learning of keypoint detector and descriptor for pose
  invariant 3d matching.
\newblock arXiv preprint (2018)

\bibitem{simo2015discriminative}
Simo-Serra, E., Trulls, E., Ferraz, L., Kokkinos, I., Fua, P., Moreno-Noguer,
  F.:
\newblock Discriminative learning of deep convolutional feature point
  descriptors.
\newblock In: ICCV. (2015)

\bibitem{tian2017l2}
Tian, B.F.Y., Wu, F.:
\newblock L2-net: Deep learning of discriminative patch descriptor in euclidean
  space.
\newblock In: CVPR. (2017)

\bibitem{han2015matchnet}
Han, X., Leung, T., Jia, Y., Sukthankar, R., Berg, A.C.:
\newblock Matchnet: Unifying feature and metric learning for patch-based
  matching.
\newblock In: CVPR. (2015)

\bibitem{yi2016lift}
Yi, K.M., Trulls, E., Lepetit, V., Fua, P.:
\newblock Lift: Learned invariant feature transform.
\newblock In: ECCV. (2016)

\bibitem{balntas2016learning}
Balntas, V., Riba, E., Ponsa, D., Mikolajczyk, K.:
\newblock Learning local feature descriptors with triplets and shallow
  convolutional neural networks.
\newblock In: BMVC. (2016)

\bibitem{wu20083d}
Wu, C., Clipp, B., Li, X., Frahm, J.M., Pollefeys, M.:
\newblock 3d model matching with viewpoint-invariant patches (vip).
\newblock In: CVPR. (2008)

\bibitem{chu2011multi}
Chu, J., Nie, C.m.:
\newblock Multi-view point clouds registration and stitching based on sift
  feature.
\newblock In: ICCRD. (2011)

\bibitem{dai2017bundlefusion}
Dai, A., Nie{\ss}ner, M., Zollh{\"o}fer, M., Izadi, S., Theobalt, C.:
\newblock Bundlefusion: Real-time globally consistent 3d reconstruction using
  on-the-fly surface reintegration.
\newblock TO \textbf{36}(3) (2017) ~24

\bibitem{endres20143}
Endres, F., Hess, J., Sturm, J., Cremers, D., Burgard, W.:
\newblock 3-d mapping with an rgb-d camera.
\newblock IEEE Transactions on Robotics \textbf{30}(1) (2014)  177--187

\bibitem{Matterport3D}
Chang, A., Dai, A., Funkhouser, T., Halber, M., Niessner, M., Savva, M., Song,
  S., Zeng, A., Zhang, Y.:
\newblock {Matterport3D}: Learning from {RGB-D} data in indoor environments.
\newblock In: 3DV. (2017)

\bibitem{handa:etal:ICRA2014}
Handa, A., Whelan, T., McDonald, J., Davison, A.:
\newblock A benchmark for {RGB-D} visual odometry, {3D} reconstruction and
  {SLAM}.
\newblock In: ICRA. (2014)

\bibitem{sturm2012benchmark}
Sturm, J., Engelhard, N., Endres, F., Burgard, W., Cremers, D.:
\newblock A benchmark for the evaluation of rgb-d slam systems.
\newblock In: IROS. (2012)

\bibitem{SSS:2006}
Snavely, N., Seitz, S.M., Szeliski, R.:
\newblock Photo tourism: Exploring photo collections in 3d.
\newblock In: SIGGRAPH. (2006)

\bibitem{su2009learning}
Su, H., Sun, M., Fei-Fei, L., Savarese, S.:
\newblock Learning a dense multi-view representation for detection, viewpoint
  classification and synthesis of object categories.
\newblock In: ICCV. (2009)

\bibitem{Su_2015_ICCV}
Su, H., Maji, S., Kalogerakis, E., Learned-Miller, E.:
\newblock Multi-view convolutional neural networks for 3d shape recognition.
\newblock In: ICCV. (2015)

\bibitem{wangdominant}
Wang, C., Pelillo, M., Siddiqi, K.:
\newblock Dominant set clustering and pooling for multi-view 3d object
  recognition.
\newblock In: BMVC. (2017)

\bibitem{qi2016volumetric}
Qi, C.R., Su, H., Nie{\ss}ner, M., Dai, A., Yan, M., Guibas, L.J.:
\newblock Volumetric and multi-view cnns for object classification on 3d data.
\newblock In: CVPR. (2016)

\bibitem{besl1992method}
Besl, P.J., McKay, N.D.,  et~al.:
\newblock A method for registration of 3-d shapes.
\newblock PAMI \textbf{14}(2) (1992)  239--256

\bibitem{pomerleau2013comparing}
Pomerleau, F., Colas, F., Siegwart, R., Magnenat, S.:
\newblock Comparing icp variants on real-world data sets.
\newblock Autonomous Robots \textbf{34}(3) (2013)

\bibitem{Rusinkiewicz:2001:EVO}
Rusinkiewicz, S., Levoy, M.:
\newblock Efficient variants of the {ICP} algorithm.
\newblock In: 3DIM. (2001)

\bibitem{yang2016go}
Yang, J., Li, H., Campbell, D., Jia, Y.:
\newblock Go-icp: a globally optimal solution to 3d icp point-set registration.
\newblock PAMI \textbf{38}(11) (2016)  2241--2254

\bibitem{briales2017convex}
Briales, J., Gonzalez-Jimenez, J.:
\newblock Convex global 3d registration with lagrangian duality.
\newblock In: CVPR. (2017)

\bibitem{strecha2008benchmarking}
Strecha, C., Von~Hansen, W., Van~Gool, L., Fua, P., Thoennessen, U.:
\newblock On benchmarking camera calibration and multi-view stereo for high
  resolution imagery.
\newblock In: CVPR. (2008)

\bibitem{chen2016multi}
Chen, X., Ma, H., Wan, J., Li, B., Xia, T.:
\newblock Multi-view 3d object detection network for autonomous driving.
\newblock arXiv preprint (2016)

\bibitem{albarelli2009matching}
Albarelli, A., Bulo, S.R., Torsello, A., Pelillo, M.:
\newblock Matching as a non-cooperative game.
\newblock In: ICCV. (2009)

\bibitem{rodola2013scale}
Rodol{\`a}, E., Albarelli, A., Bergamasco, F., Torsello, A.:
\newblock A scale independent selection process for 3d object recognition in
  cluttered scenes.
\newblock IJCV \textbf{102}(1-3) (2013)

\bibitem{ijcai2017-627}
Jiayi, M., Ji, Z., Hanqi, G., Junjun, J., Huabing, Z., Yuan, G.:
\newblock Locality preserving matching.
\newblock In: IJCAI. (2017)

\bibitem{black1996unification}
Black, M.J., Rangarajan, A.:
\newblock On the unification of line processes, outlier rejection, and robust
  statistics with applications in early vision.
\newblock IJCV \textbf{19}(1) (1996)  57--91

\bibitem{Han_2015_CVPR}
Han, X., Leung, T., Jia, Y., Sukthankar, R., Berg, A.C.:
\newblock Matchnet: Unifying feature and metric learning for patch-based
  matching.
\newblock In: CVPR. (2015)

\bibitem{HardNet2017}
Anastasiya, M., Dmytro, M., Filip, R., Jiri, M.:
\newblock Working hard to know your neighbor's margins: Local descriptor
  learning loss.
\newblock In: NIPS. (2017)

\bibitem{johnson2016perceptual}
Johnson, J., Alahi, A., Fei-Fei, L.:
\newblock Perceptual losses for real-time style transfer and super-resolution.
\newblock In: ECCV. (2016)

\bibitem{he2016deep}
He, K., Zhang, X., Ren, S., Sun, J.:
\newblock Deep residual learning for image recognition.
\newblock In: CVPR. (2016)

\bibitem{szegedy2015going}
Szegedy, C., Liu, W., Jia, Y., Sermanet, P., Reed, S., Anguelov, D., Erhan, D.,
  Vanhoucke, V., Rabinovich, A.:
\newblock Going deeper with convolutions.
\newblock In: CVPR. (2015)

\bibitem{szegedy2016rethinking}
Szegedy, C., Vanhoucke, V., Ioffe, S., Shlens, J., Wojna, Z.:
\newblock Rethinking the inception architecture for computer vision.
\newblock In: CVPR. (2016)

\bibitem{lin2013network}
Lin, M., Chen, Q., Yan, S.:
\newblock Network in network.
\newblock arXiv preprint (2013)

\bibitem{2015arXiv150500853X}
{Xu}, B., {Wang}, N., {Chen}, T., {Li}, M.:
\newblock Empirical evaluation of rectified activations in convolutional
  network.
\newblock arXiv preprint (2015)

\bibitem{chopra2005learning}
Chopra, S., Hadsell, R., LeCun, Y.:
\newblock Learning a similarity metric discriminatively, with application to
  face verification.
\newblock In: CVPR. (2005)

\bibitem{lin2017deephash}
Lin, J., Mor{\`e}re, O., Veillard, A., Duan, L.Y., Goh, H., Chandrasekhar, V.:
\newblock Deephash for image instance retrieval: Getting regularization, depth
  and fine-tuning right.
\newblock In: ICMR. (2017)

\bibitem{snavely2008modeling}
Snavely, N., Seitz, S.M., Szeliski, R.:
\newblock Modeling the world from internet photo collections.
\newblock IJCV \textbf{80}(2) (2008)  189--210

\bibitem{goesele2007multi}
Goesele, M., Snavely, N., Curless, B., Hoppe, H., Seitz, S.M.:
\newblock Multi-view stereo for community photo collections.
\newblock In: ICCV. (2007)

\bibitem{Zhou_2017_ICCV}
Zhou, L., Zhu, S., Shen, T., Wang, J., Fang, T., Quan, L.:
\newblock Progressive large scale-invariant image matching in scale space.
\newblock In: ICCV. (2017)

\bibitem{shen2016graph}
Shen, T., Zhu, S., Fang, T., Zhang, R., Quan, L.:
\newblock Graph-based consistent matching for structure-from-motion.
\newblock In: ECCV. (2016)

\bibitem{zhu2017parallel}
Zhu, S., Shen, T., Zhou, L., Zhang, R., Wang, J., Fang, T., Quan, L.:
\newblock Parallel structure from motion from local increment to global
  averaging.
\newblock arXiv preprint arXiv:1702.08601 (2017)

\bibitem{Zhang_2017_ICCV}
Zhang, R., Zhu, S., Fang, T., Quan, L.:
\newblock Distributed very large scale bundle adjustment by global camera
  consensus.
\newblock In: ICCV. (2017)

\bibitem{li2016efficient}
Li, S., Siu, S.Y., Fang, T., Quan, L.:
\newblock Efficient multi-view surface refinement with adaptive resolution
  control.
\newblock In: ECCV. (2016)

\bibitem{raguram2008comparative}
Raguram, R., Frahm, J.M., Pollefeys, M.:
\newblock A comparative analysis of ransac techniques leading to adaptive
  real-time random sample consensus.
\newblock In: ECCV. (2008)

\bibitem{cooper1990computational}
Cooper, G.F.:
\newblock The computational complexity of probabilistic inference using
  bayesian belief networks.
\newblock Artificial intelligence \textbf{42}(2-3) (1990)  393--405

\bibitem{murphy1999loopy}
Murphy, K.P., Weiss, Y., Jordan, M.I.:
\newblock Loopy belief propagation for approximate inference: An empirical
  study.
\newblock In: UAI. (1999)

\bibitem{tatikonda2002loopy}
Tatikonda, S.C., Jordan, M.I.:
\newblock Loopy belief propagation and gibbs measures.
\newblock In: UAI. (2002)

\bibitem{yedidia2003understanding}
Yedidia, J.S., Freeman, W.T., Weiss, Y.:
\newblock Understanding belief propagation and its generalizations.
\newblock Exploring artificial intelligence in the new millennium (2003)
  239--269

\bibitem{openblas}
Zhang, X., Wang, Q., Werner, S., Zaheer, C., Chen, S., Luo, W.:
\newblock http://www.openblas.net/

\bibitem{Rusu_ICRA2011_PCL}
Rusu, R.B., Cousins, S.:
\newblock 3d is here: Point cloud library (pcl).
\newblock In: ICRA. (2011)

\bibitem{hpatches_2017_cvpr}
Balntas, V., Lenc, K., Vedaldi, A., Mikolajczyk, K.:
\newblock Hpatches: A benchmark and evaluation of handcrafted and learned local
  descriptors.
\newblock In: CVPR. (2017)

\bibitem{huang2018learning}
Huang, H., Kalogerakis, E., Chaudhuri, S., Ceylan, D., Kim, V.G., Yumer, E.:
\newblock Learning local shape descriptors from part correspondences with
  multiview convolutional networks.
\newblock TOG \textbf{37}(1) (2018) ~6

\bibitem{flint2007thrift}
Flint, A., Dick, A., Van Den~Hengel, A.:
\newblock Thrift: Local 3d structure recognition.
\newblock In: Digital Image Computing Techniques and Applications. (2007)

\bibitem{schoenberger2016mvs}
Sch\"{o}nberger, J.L., Zheng, E., Pollefeys, M., Frahm, J.M.:
\newblock Pixelwise view selection for unstructured multi-view stereo.
\newblock In: ECCV. (2016)

\bibitem{dai2017scannet}
Dai, A., Chang, A.X., Savva, M., Halber, M., Funkhouser, T., Nie{\ss}ner, M.:
\newblock Scannet: Richly-annotated 3d reconstructions of indoor scenes.
\newblock In: CVPR. (2017)

\bibitem{kerl2013robust}
Kerl, C., Sturm, J., Cremers, D.:
\newblock Robust odometry estimation for rgb-d cameras.
\newblock In: ICRA. (2013)

\bibitem{Choi_2015_CVPR}
Choi, S., Zhou, Q.Y., Koltun, V.:
\newblock Robust reconstruction of indoor scenes.
\newblock In: CVPR. (2015)

\bibitem{wald2001interactive}
Wald, I., Slusallek, P., Benthin, C., Wagner, M.:
\newblock Interactive rendering with coherent ray tracing.
\newblock In: Computer graphics forum. (2001)

\end{thebibliography}
\end{document}